\documentclass[12pt]{article}   	
\usepackage{geometry}                		
\usepackage[noframe]{showframe}
\usepackage{graphicx}				
\usepackage{amssymb}
\usepackage{amsmath}
\usepackage{amsthm}
\usepackage[]{algorithm2e}
\usepackage{array}
\usepackage{mathtools}
\allowdisplaybreaks

\usepackage{afterpage}
\usepackage{threeparttable}
\usepackage{tikz}
\usepackage{emptypage}

\usepackage{longtable}
\usepackage{booktabs}
\newcommand{\ra}[1]{\renewcommand{\arraystretch}{#1}}
\usepackage{arydshln}
\usepackage{makecell}
\usepackage{pbox}
\usepackage{epstopdf}

\DeclareMathOperator*{\argmin} {argmin}

\usepackage{subcaption}
\usepackage{parskip}
\usepackage{array}
\newcolumntype{P}[1]{>{\centering\arraybackslash}p{#1}}
\newcolumntype{M}[1]{>{\centering\arraybackslash}m{#1}}
\usepackage{multirow}
\usepackage{bm}

\newcommand{\norm}[1]{\left\lVert#1\right\rVert}

\usepackage[utf8]{inputenc}
\usepackage[english]{babel}

\usepackage{natbib}
\setcitestyle{authoryear,open={(},close={)}}

\usepackage{csquotes}
\renewcommand{\mkbegdispquote}[2]{\itshape}

\usepackage{authblk}

\usepackage[toc, page]{appendix}

\usepackage{bbm}
\usepackage{fancyhdr}
\title{Concentrated Document Topic Model}
\author[1]{Hao LEI }
\author[2,3]{Ying CHEN }
\affil[1]{Department of Statistics and Applied Probability, National University of Singapore}
\affil[2]{Department of Mathematics, National University of Singapore}
\affil[3]{Risk Management Institute, National University of Singapore}

\date{}
\begin{document}
\maketitle
\begin{abstract}
	We propose a Concentrated Document Topic Model(CDTM) for unsupervised text classification, which is able to produce a concentrated and sparse document topic distribution. In particular, an exponential entropy penalty is imposed on the document topic distribution. Documents that have diverse topic distributions are penalized more, while those having concentrated topics are penalized less. We apply the model to the benchmark NIPS dataset and observe more coherent topics and more concentrated and sparse document-topic distributions than Latent Dirichlet Allocation(LDA).
\end{abstract}

\section{Introduction}

Probabilistic topic modeling is a popular method to cluster data into different groups and reduce the dimension. It has been applied in many different areas \citep{blei2007correlated, huang2017analyst, maier2018applying, reisenbichler2019topic,lei2020investor, FEI, liu2016overview,gonzalez2019cistopic}. Much attention in previous studies are focused on improving the estimated topic quality \cite{griffiths2005integrating, wallach2009rethinking,das2015gaussian, shi2017jointly, xu2018distilled}. And little is paid on the document-topic distributions. Yet in real world applications, the document-topic distributions are no less, if not more, important than the topic contents, i.e. the topic-word distributions. For example, in an information retrieval  task, the document-topic distributions decide how accurately the revelant documents can be retrieved for a given topic. Existing probabilistic topic models make little assumptions on the document-topic distribution and rely on the posterior maximization. For example, Latent Dirichlet Allocation assumes that the document-topic distribution is drawn from a Dirichlet distribution and topics are drawn from it following a multinormial distributions. As a result of the minimal number of assumptions, the estimated document-topic distribution can take any form, as long as they are proper multinomial parameters, i.e. elements are non-negative and sum up to 1. One extreme case is that a document contains significant portions of all topics. This does not inline with the reality, in which documents only contain a few topics. In the information retrieval task, this document will not be accurately mapped to the users' query. To fill this gap, we propose a Concentrated Document Topic Model (CDTM). Our proposed CDTM is able to produce concentrated and sparse document-topic distributions. In particular, we add the entropies of the document-topic distributions to the model posterior, which encourages low entropy values, thus encouraging concentrated and sparse document-topic distributions.

Two earliest work in the field is Probabilistic Latent Semantic Indexing(PSLI) \citep{PLSI} and Latent Dirichlet Allocation (LDA) \citep{LDA}. One difference of the two models is that PSLI assumes that every document has only one topic, while LDA allows multiple topics in a document. LDA assumes the document-topic parameter is drawn from a Dirichlet prior. Thus the topics are almost independent of each other. \cite{blei2007correlated} extend LDA to a correlated topic model. Instead of the Dirichlet distribution, it assumes the document-topic parameter is from a logistic normal distribution. The covariance matrix is used to model the topic correlation inside the documents. Their empirical experiment shows that the correlated topic model fits better than the LDA and supports more topics than LDA, i.e. the log probability of LDA on test dataset peaks near 30 topics, while that of the correlated topic model is 90 topics.  \cite{wallach2009rethinking} investigate the effect of asymmetric priors on the performance of LDA. The evaluation metric is the log-probability per word in the training dataset and probability on the test dataset. Their study shows that the combination of asymmetric prior on document-topic parameters and symmetric prior on topic-words parameters produces the best results on several benchmark dataset. They also provide a possible explanation of the superior performance of the combination. Namely, the asymmetric prior on document-topic parameters serves to share the common words across different documents and the symmetric prior on topic-word parameters serves to distinguish different topics. Their study in some way supports our proposed method.

Our proposed CDTM make use of entropy \citep{shannon1948mathematical} to measure the topic concentration. Entropy is widely used in estimating the functional form of density. \cite{zellner1988calculation} propose to estimate the density by maximizing the entropy while subjecting to moments constraints. They show that the estimates takes an exponential polynomial form where the coefficients are numerically computed using Taylor series expansion and Newton's method. \cite{ryu1993maximum} extends the Maximum Entropy(ME) density \citep{zellner1988calculation}  to a flexible ME density by replacing the moments constraints with constraints on known functions and the ME regression functions by replacing the density with the regression function. They show that several well-known econometric functions, e.g. exponential polyponomial, Cobb-Douglas, translog, generalized Leontief, Fourier flexible form can be derived using this approach.
\cite{park2009maximum} combines the ME density estimator with ARCH series models \citep{engle1982autoregressive, bollerslev1986generalized}. The moment constraints are extended to constraints with additional parameters. Several moment functions are used to capture excess kurtosis, asymmetry and high peakedness in financial data. Entropy has also been used as a penalization term. \cite{gomes2007entropy} use the entropy penalty as a regularization when estimating the viscosity solution of the Hamilton-Jacobi equation. \cite{koltchinskii2009sparse} theoretically study the sparsity estimation in convex hulls using entropy penalization. They show that the 'approximate sparsity' of the solution to the theoretical risk minimization problem under the entropy penalization implies the `approximate sparsity` of its counterparty in the empirical risk minimization problem. They also explore various bounds on the excess risk of the empirical solution. A similar idea is explored by \cite{koltchinskii2011neumann} when estimating low-rank matrix. Instead of Shannon entropy, they use the von Neumann entropy as the penalty term. They show that when the target matrix is nearly low-rank, the empirical estimator is well approximated by low-rank matrices and $L_2$ error can be controlled in terms of the 'approximate rank' of the target matrix. Entropy penalty is also widely used in deep learning and reinforcement learing. \cite{williams1991function} and \cite{mnih2016asynchronous} find that penalizing the low entropy of the policy improves the reinforcement learning exploration by discouraging premature convergence to suboptimal deterministic policies. The entropy penalty is also used by \cite{luo2017learning} to train an online sequence-to-sequence model. The entropy of the emission policy was added in the objective function. \cite{pereyra2017regularizing} add the entropy of the output to the objective function. They test the proposed method on six common benchmarks, e.g. image classification, machine translation, etc and find that the penalty improves the state-of-art models across benchmarks without modifying existing hyperparameters.

The rest of this paper is organized in the following way. In section \ref{sec:slda}, we provide details of the proposed method and the algorithms to estimate the parameters.  In section \ref{sec:nips}, we apply the proposed method to the public NIPS dataset. The results show that the proposed method improves the topic coherence and encourages concentration and sparsity on the document-topic distributions. We conclude the paper in section \ref{sec:conclusion}.

\section{Method} \label{sec:slda}

The model set up is the same as LDA \citep{LDA}. Namely, given a corpus $C$, we assume it contains $K$ topics. Every topic $\eta_k$ is multinomial distribution on the vocabulary. Every document $d$ contains one or more topics. The topic proportion in each document is governed by the local latent parameter document-topic $\theta$, which has a Dirichlet prior with hyperparameter $\zeta$.  Every word in document $d$ is generated from the contained topics as follows:

\begin{itemize}
	\item for every document $d \in C$, its topic proportion parameter $\theta$ is generated from a Dirichlet distribution, i.e. $\theta \sim Dir(\zeta)$.
	\item for every word in the document $d$,
	\begin{itemize}
		\item a topic $Z$ is first generated from the multinomial distribution with parameter $\theta$, i.e. $Z \sim Multinomial(\theta)$
		\item a word $w$ is then generated from the multinomial distribution with parameter $\eta_Z$, i.e. $w \sim Multinomial(\eta_Z)$
	\end{itemize}
\end{itemize}

 \begin{figure}[htbp]
	\begin{center}
		\includegraphics[width=\textwidth, height=0.4\textheight]{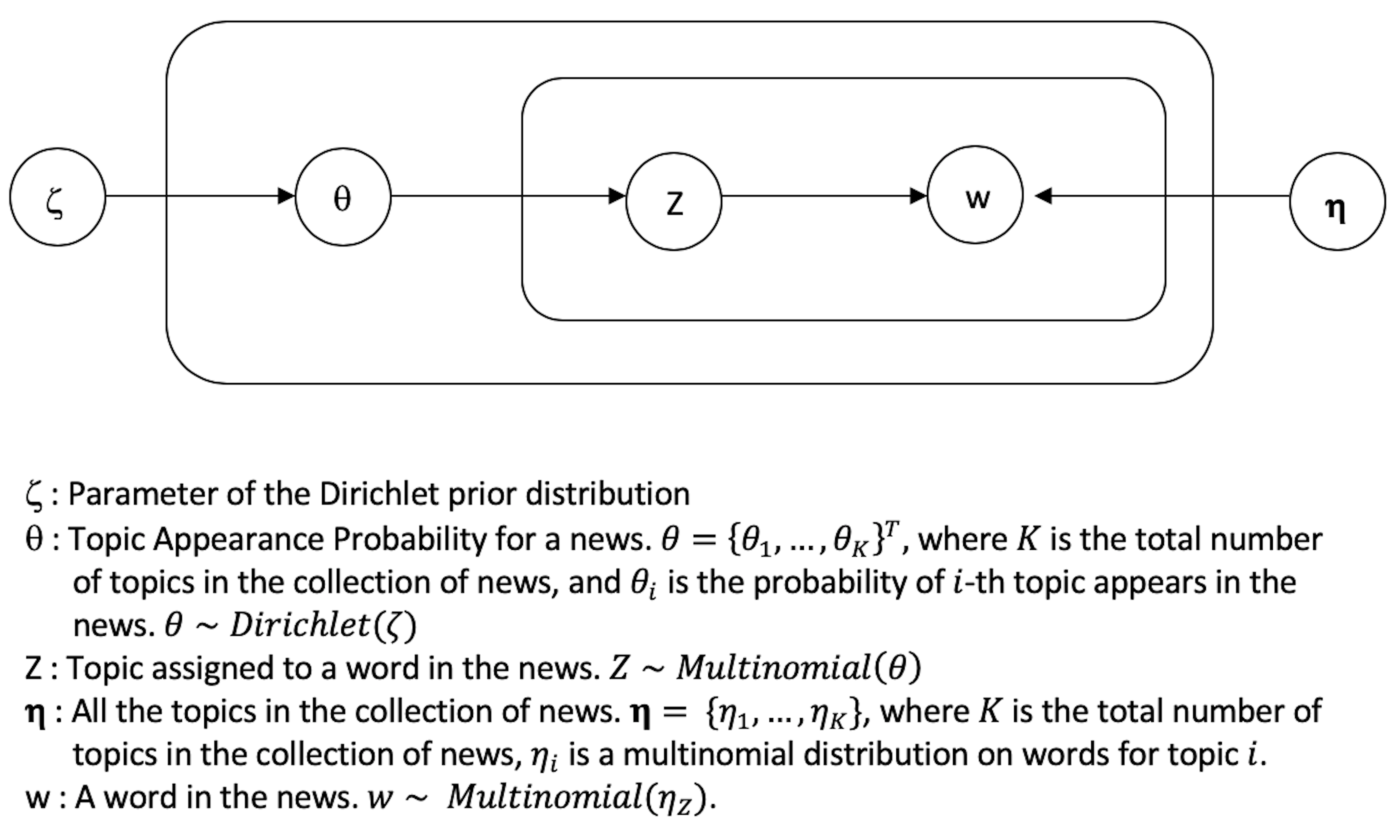}
		\caption{ The graphical representation. The outer box represents the document level. The inner rectangle represents the word level. }
		\label{fig:lda}
	\end{center}
\end{figure}

The graphical representation is shown in Figure \ref{fig:lda}. The outer rectangle represents the document-level and the inner rectangle represents the word-level. $\zeta$ and $\eta$ are global parameters, i.e. shared by all the documents. $\theta$ and $Z$ are local latent variables. A complete Bayesian approach further assumes that topics $\eta_1, \dots, \eta_K$  are generated from a Dirichlet prior with hyperparameter $\beta$. Here we use this formulation as in \cite{LDA} for the ease of adding a penalty.

The latent parameters in CDTM are estimated by maximizing the following penalized posterior.
\begin{equation}
	\max \hspace{2mm}  p(\bm{\theta}, \bm{Z},  \zeta, \eta | W )\prod_{d=1}^{D}\exp(-\lambda_d H(\theta_d))
\end{equation}
where $D$ is the number of documents in the corpus, $H(\theta_d) = -\sum_{i=1}^{K}\theta_{di}\log \theta_{di}$ is the entropy of document-topic distribution for document $d$, and 
\begin{equation}\label{equ:posterior}
	p(\bm{\theta}, \bm{Z},  \zeta, \eta | W ) = \prod_{d=1}^{D} p(\theta_d|\zeta) \prod_{n=1}^{N_d} p(Z_{dn}|\theta_d)p(w_{dn}|Z_{dn}, \eta)
\end{equation}
is the posterior and $N_d$ is the number of words for document $d$. In the model, documents are assumed to be exchangable. In the following discussion, we omit the subscript $d$ and limit the discussion for a single document. When a document only contains a couple of concentrated topics, the entropy  $H(\theta)$ would be small and penalty is close to $1$. On contrary, when the document topics are equally distributed across all the topics, the entropy would achieve its largest value and the penalty value is closer to $0$. Therefore, this productive penalty encourages concentrated and sparsity in document-topic distributions and penalizes diverse and equally distributed document-topic parameters.

The posterior $ p(\theta, Z,  \zeta, \eta | W )$ is intractable. We maximize the Evidence Lower Bound (ELBO) \citep{LDA, blei2017variational}
 $L( \zeta, \eta, \gamma, \phi)$,
 \begin{align*}
 	 L( \zeta, \eta, \gamma, \phi) &=  E_q[ \ln p(\theta, Z, W | \zeta, \eta) - \lambda H(\theta)]-E_q[\ln q(\theta,Z| \gamma, \phi) ] \\
 	 & =  E_q[ \ln p(\theta, Z, W | \zeta, \eta)]-E_q[\ln q(\theta,Z| \gamma, \phi) ] + E_q [- \lambda H(\theta)]
 \end{align*}
where $p(.)$ is the single document version of equation \ref{equ:posterior} and $q(\theta, Z | \gamma, \phi)$ is the mean-field variational distribution
\begin{equation*}
	q(\theta, Z | \gamma, \phi) = q(\theta | \gamma)\prod_{n=1}^{N} q(Z_n|\phi_n)
\end{equation*}
where $N$ is the number of words in a document, $q(\theta | \gamma) \sim Dirichlet(\gamma)$, and $q(Z_n|\phi_n) \sim Multinomial(\phi_n)$. $E_q$ represents the expectation under the variational distribution. The variational distribution decouples the $\theta$ and $Z_n$ and simplifies the intractable computation.

The ELBO $L(.)$ is maximized in an 'EM'-like steps. In the E-step, we maximize the ELBO w.r.t. the local latent parameters $\gamma, \phi$ for every document, while conditioning on the global latent parameters $\zeta, \eta$. In the M-step, we maximize the ELBO w.r.t. to global latent parameters  $\zeta, \eta$ while conditioning on the local latent parameters $\gamma, \phi$. Since the penalty is a function of document-topic distribution $\theta$, it doesn't affect the estimation of $\phi$, which will be the same as that in \cite{LDA},
\begin{equation*} \label{equ:e-equ}
	\phi_{ni} \propto \eta_{iw_n} \exp{E_q[\log (\theta_i) | \gamma]} \\
\end{equation*}
where
\begin{equation*}
	\exp{E_q[\log (\theta_i) | \gamma]} = \Psi(\gamma_i) - \Psi(\sum_{l=1}^K \gamma_l)
\end{equation*}
where  $\Psi$ is the digamma function.

Taking out all the terms containing $\gamma$ from the ELBO, we have the following function
\begin{equation} \label{equ:elbo_gamma}
	\begin{aligned}
	L_{[\gamma]} = & \textstyle \sum_{i=1}^{K} (\Psi(\gamma_i) - \Psi(\sum_{l=1}^K \gamma_l))(\zeta_i + \sum_{n=1}^{N}\phi_{ni} - \gamma_i) - \log \Gamma(\sum_{l=1}^{K} \gamma_l) + \sum_{i=1}^{K}\log \Gamma(\gamma_i)+\\
		& \lambda\Bigg(\textstyle \sum_{i=1}^{K}\gamma_l \Psi(\gamma_l)/\sum_{l=1}^{K}\gamma_l -\Psi(\sum_{l=1}^{K}\gamma_l) + (K-1)/ \sum_{l=1}^{K}\gamma_l\Bigg)
	\end{aligned}
\end{equation}
where $\Gamma, \Psi$ are the gamma and digamma function, $K$ is the number of topics, $N$ is the number of words in the document. The last term is $E_q [- \lambda H(\theta)]$, which is the expected entropy under the variational distribution (see appendix for the derivation) and not separable.  There is no closed-form solution for equation \ref{equ:elbo_gamma}. We use the coordinate descent algorithm to estimate the document-topic distribution parameter $\gamma$ for every document.
\begin{equation}
	\gamma_i^{s+1} \coloneqq \argmin -L_{[\gamma]}(\gamma_1^{s}, \dots, \gamma_{i-1}^s, x, \gamma_{i+1}^s, \dots, \gamma_{K}^s), i=1, \dots, K
\end{equation}
The minimization can be solved using Newton's method.
\begin{equation}
	\gamma_i^{t+1} \coloneqq \gamma_i^{t} - \alpha \frac{L^\prime_{\gamma_i}}{L^{\prime\prime}_{\gamma_i}}
\end{equation}
where
\begin{equation*}
	\begin{aligned}
		L^\prime_{\gamma_i} = &\textstyle \Psi^\prime(\gamma_i)(\zeta_i + \sum_{n=1}^N \phi_{ni} - \gamma_i) - \Psi^\prime(\sum_{l=1}^{K}\gamma_l) \sum_{l=1}^{K}(\zeta_l + \sum_{n=1}^{N}\phi_{nl} - \gamma_l) +\\
		& \textstyle \lambda \Big(\big(\Psi(\gamma_i) + \gamma_i\Psi^\prime(\gamma_i)\big)/ \sum_{l=1}^{K}\gamma_l - \sum_{l=1}^{K} \gamma_l \Psi(\gamma_l)/(\sum_{l=1}^{K}\gamma_l)^2 - \Psi^\prime(\sum_{l=1}^{K} \gamma_l) - (K-1)/(\sum_{l=1}^{K} \gamma_l)^2\Big)
	\end{aligned}
\end{equation*}
is the partial derivative of $L_{[\gamma]}$ with respect to $\gamma_i$ and
\begin{equation*}
	\begin{aligned}
		L^{\prime\prime}_{\gamma_i} = &\textstyle \Psi^{\prime\prime}(\gamma_i)(\zeta_i + \sum_{n=1}^N \phi_{ni} - \gamma_i) - \Psi^\prime(\gamma_i) - \Psi^{\prime\prime}(\sum_{l=1}^{K}\gamma_l) \sum_{l=1}^{K}(\zeta_l + \sum_{n=1}^{N}\phi_{nl} - \gamma_l) + \Psi^\prime(\sum_{l=1}^{K} \gamma_l) +\\
		& \textstyle \lambda \Big(\big(2\Psi^\prime(\gamma_i) + \gamma_i\Psi^{\prime\prime}(\gamma_i)\big)/ \sum_{l=1}^{K}\gamma_l -
		2(\Psi(\gamma_i) + \gamma_i\Psi^\prime(\gamma_i)) / (\sum_{l=1}^{K}\gamma_l)^2 + \\
		&\hspace*{4mm} \textstyle 2(K-1 + \sum_{l=1}^{K}\gamma_i\Psi(\gamma_i))/ (\sum_{l=1}^{K}\gamma_l)^3 - \Psi^{\prime\prime}(\sum_{l=1}^{K} \gamma_l)\Big)
	\end{aligned}
\end{equation*}
is the second partial derivative of $L_{[\gamma]}$ with respect to $\gamma_i$. The step $\alpha$ can be found using backtracking line search. The complete algorithm for the e-step for a document is given in Algorithm \ref{e_algo}.

In the M-step, we maximize the ELBO with respect to $\eta$ and the updating equation is
\begin{equation*}
	\eta_{ij} \propto \sum_{d=1}^{D}\sum_{n=1}^{N_d} \phi_{dni}w_{dn}^j
\end{equation*}

\begin{algorithm}[H] \label{e_algo}
	\KwResult{Update the words topic assignment parameter $\phi$ and the document topic parameter $\gamma$ for a document $d$}
	Initialize the $\gamma_i = \alpha_i + N/K, \forall i=1, \dots, K$ and $\phi_{ni}^0 = 1/K, \forall n=1,\dots, N, i=1,\dots,K$\;
	Choose the stopping criterion $\epsilon$ and line searching parameter $\alpha \in (0, 0.5)$, $\rho \in (0, 1)$ \;
	\While{Not converge}{
		\For{word $n, n=1, \dots, N$}{
			\For{document $i, i=1, \dots, K$}{
				$\phi_{ni}^{t+1} \coloneqq \eta_{iw_n}\exp{E_q[\log (\theta_i) | \gamma]}$
				}
			Normalize $\phi_n^{t+1}$ so that its sum equals to 1.
			}
		\For{topic $i, i=1, \dots, K$}{
			\While{Not converge}{
				Compute the Newton step $\Delta \gamma_i$:
				$\Delta \gamma_i \coloneqq -L^\prime_{\gamma_i}/L^{\prime\prime}_{\gamma_i}$\;
				\eIf {$|\Delta \gamma_i| < \epsilon$}{Stop updating $\gamma_i$}{
					Find step size $\alpha$ by backtracking line search: \newline
						Initialize the step size $\alpha \coloneqq 1$\;
						\While{$-L(\gamma_i + \alpha \Delta \gamma_i) > -L(\gamma_i) - \delta \alpha L^\prime_{\gamma_i} \Delta \gamma_i $}{
						\hspace*{4mm}$\alpha \coloneqq \rho \alpha$
					}
				}
			$\gamma_i \coloneqq \gamma_i + \alpha \Delta \gamma_i$\;
		}
	}
}
	\caption{Variational E-step}
\end{algorithm}

\section{Real Data Application} \label{sec:nips}
To test the empirical performance of our proposed method, we apply LDA and CDTM to the NIPS dataset, which consists of 11,463 words and 7,241 NIPS conference paper from 1987 to 2017. The data is randomly split into two parts: training (80\%) and testing (20\%). We select the number of topics being $10$ from $\{5, 10, 15, 20, 25, 30\}$ for LDA using cross-validation with perplexity as the evaluation measure on the training dataset \citep{LDA}. For CDTM, we use homogeneous hyperparameters in the experiment, i.e. $\lambda_d = \lambda, \forall d \in \{1, \dots, D\}$. The model is flexibel to use any other weighting schemes, such as a series of increasing $\lambda_d$s for the increasing entropy values. 

We use cross-validation to select the penalty weight $\lambda = 35$ from $\{25, 30, 35, 40, 45\}$. The evaluation metric is the topic coherence score $C_V$ \citep{roder2015exploring}, which has been shown achieving the highest correlation with all available human topic ranking data \citep{roder2015exploring, syed2017full}. The $C_V$ coherence score is calculated as follows. The top $N$ words of each topic are selected as the representation of the topic, denoted as $W = \{w_1, \dots, w_N\}$. Each word $w_i$ is represented by an $N$-dimensional vector $v(w_i) = \{NPMI(w_i, w_j) \}_{j=1, \dots, N}$, where $j$th-entry is the Normalized Pointwise Mutual Information(NPMI) between word $w_i$ and $w_j$, i.e. $NPMI(w_i, w_j) = \frac{\log P(w_i, w_j)  - \log (P(w_i)P(w_j))}{-\log P(w_i, w_j)}$. $W$ is represented by the sum of all word vectors, $v(W) = \sum_{j=1}^N v(w_j)$. The calculation of NPMI between word $w_i$ and $w_j$ involves the marginal and joint probabilities $p(w_i), p(w_j), p(w_i, w_j)$. A sliding window of size 110, which is the default value in the python package 'gensim' and robust for many applications,  is used to create pseudo-document and estimate the probabilities. The purpose of the sliding window is to take the distance between two words into consideration. For each word $w_i$, a pair is formed $(v(w_i), v(W))$. A cosine similarity measure $\phi_i(v(w_i), v(W)) = \frac{v(w_i)^T v(W)}{\norm{v(w_i)} \norm{v(W)} }$ is then calculated for each pair. The final $C_V$ score for the topic is the average of all $\phi_i$s.

\begin{figure}
	\begin{center}
		\includegraphics[width=0.8\textwidth, height=8cm]{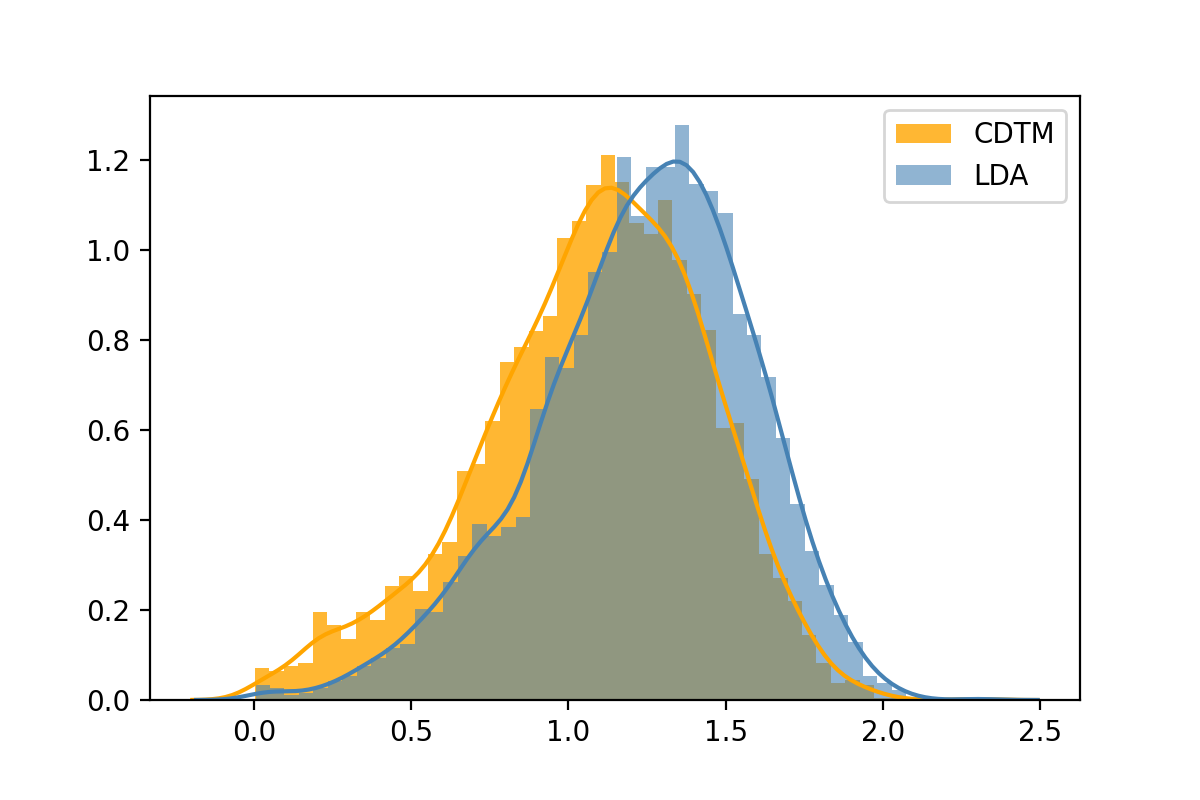}
		\caption{Histogram of the entropies of document-topic distributions.}
		\label{fig:entropy_histogram}
	\end{center}
\end{figure}

We then refit both LDA and CDTM to the whole training dataset and use the test dataset to calculate the coherence score $C_V$ as a final evaluation of the out-of-sample performance. The results are shown in Table \ref{tab:top20}. Overall the $C_V$ score of LDA topics is 0.50 and that of CDTM is 0.52, a $5\%$ improvement. We then compute the entropy of the document-topic distributions and plot its histrogram in figure \ref{fig:entropy_histogram}. The summary statistics of the entropy values is shown in Table \ref{tab:summary_stats}. From both the plot and summary statistics, we observe that the entropy of the document-topic distributions of CDTM is on average smaller (1.08) than that of LDA (1.24), indicating that overall the document-topic is more concentrated in CDTM than LDA. 

We present a specific example: the paper titled \textit{`Connection Topology and Dynamics in Lateral Inhibition Networks'} \citep{marcus1990connection}. This paper studies dynamics of modeling the lateral inhibition, a topic in neurobiology. A paragraph from the paper is quoted below. 
 \begin{displayquote}
In this paper we study the dynamics of simple neural network models of lateral inhibition
in a variety of two-dimensional connection schemes. The lattice structures we study are
shown in Fig. 1. Two-dimensional lattices are of particular importance to artificial
vision systems because they allow an efficient mapping of an image onto a network and
because they are well-suited for implementation in VLSI circuitry. We show that the
stability of these networks depends sensitively on such design considerations as local
connection topology, neuron self-coupling, the steepness or gain of the neuron transfer
function, and details of the network dynamics such as connection delays for continuoustime dynamics or update rule for discrete-time dynamics.   
\end{displayquote}
LDA assigns two significant topics to this paper: $0.39$ to topic \textit{Neural Network} and $0.42$ to topic \textit{Neurobiology}; while CDTM assign $0.93$ to topic \textit{Neurobiology}. Although this paper have words \textit{neural network}, it has nothing to do with the modern computer science neural network with hidden layers, training etc. In fact, the paper doesn't even contain words like \textit{learning, layer, training, hidden, error, weight} etc. LDA wrongly assigns $0.39$ to topic \textit{Neural Network}. Moreover, the sum of the two main topic probabilities in LDA is $0.81$, which is smaller than the $0.93$ assigned by CDTM. 

\begin{table}[htbp]
	\centering
	\ra{1.1}
	
	\begin{tabular*}{\columnwidth} {@{\extracolsep{\fill}} lrrrr}
		\toprule
		& Mean & Variance & Skewness & Kurtosis \\
		\midrule
		LDA  & 1.24 &	0.12&	-0.48&	0.17 \\
		CDTM & 1.08 &	0.13&	-0.43&	0.01 \\
		\bottomrule
	\end{tabular*}
   \caption{Summary statistics of the entropies of document-topic distributions.}
	\label{tab:summary_stats}
\end{table}

\begin{table}[htbp]
	\centering
	\ra{1.3}
	\footnotesize
	\begin{tabular*}{\columnwidth} {@{\extracolsep{\fill}}p{2cm} p{13cm}r}
		\toprule
		LDA & & 0.50\\
		\midrule
Neural Network &		network   networks   neural   input   learning   layer   output   training   units   hidden   weights   error   time   function   weight   set    figure   used   number   using        & 0.52 \\
Optimization &		algorithm   bound   theorem   function   learning   let    log    loss   set    bounds   case   functions   algorithms   error   probability   problem   convex   proof   following   lemma        & 0.42 \\
Reinforcement Learning &		state   learning   policy   time   action   value   reward   function   algorithm   optimal   agent   actions   states   problem   reinforcement   decision   model   control   using   based        & 0.56 \\
Statistics &		model   data   distribution   gaussian   models   log    likelihood   using   posterior   prior   parameters   function   mean   bayesian   density   distributions   estimation   process   estimate   sample        & 0.46 \\
Classification &		learning   training   data   classification   model   features   feature   using   task   class   set    test   classifier   deep   dataset   performance   models   label   use    used        & 0.53 \\
Machine learning &		model   models   graph   tree   variables   inference   node   nodes   set    structure   algorithm   number   markov   probability   time   topic   distribution   given   variable   figure        & 0.38 \\
Neurobiology &		model   time   neurons   neuron   figure   neural   spike   response   information   input   activity   stimulus   cells   cell   signal   fig    brain   synaptic   different   rate        & 0.63 \\
Computer Vision &		image   images   object   model   figure   features   using   objects   visual   recognition   human   feature   vision   detection   segmentation   motion   set    based   used   different        & 0.59 \\
Statistical learning &		matrix   kernel   problem   algorithm   linear   method   sparse   methods   data   optimization   rank   solution   norm   vector   matrices   analysis   using   convex   function   error        & 0.44 \\
Clustering &		data    algorithm   clustering   set    number   points   graph   distance   cluster   clusters   problem   algorithms   based   time   using   means   information   query   point   figure        & 0.43 \\
		\midrule
		CDTM & & 0.52 \\
		\midrule         
Neural Network	&	network   networks   neural   training   learning   layer   input   output   hidden   units   deep   model   using   trained   used   weights   error   set    recognition   layers        & 0.61 \\

Optimization &		algorithm   optimization   problem   function   gradient   convex   algorithms   method   solution   set    methods   convergence   learning   problems   linear   loss   functions   time   objective   stochastic        & 0.52 \\
Reinforcement learning &		learning   state   policy   algorithm   action   time   value   function   reward   regret   optimal   agent   problem   set    actions   reinforcement   states   decision   probability   using         & 0.55 \\
Statistics &		model   data   distribution   models   gaussian   log    likelihood   inference   parameters   posterior   bayesian   using   prior   latent   process   time   sampling   variables   mean   variational   &      0.51 \\
Classification &		learning   data   training   classification   kernel   set    features   feature   class   using   test   classifier   label   examples   used   based   number   labels   performance   task     &    0.55 \\
Machine learning &		graph   set    tree   algorithm   node   nodes   clustering   model   number   structure   variables   cluster   graphs   models   given   clusters   time   data   random   probability        & 0.47 \\
Neurobiology &		model   figure   visual   response   stimulus   human   brain   cells   information   data   different   responses   spatial   stimuli   target   subjects   time   motion   cell   task        & 0.59 \\
Computer Vision &		image   images   object   model   using   features   figure   feature   recognition   objects   vision   set    detection   visual   segmentation   based   use    different   local   used        & 0.55 \\
Statistical learning &		matrix   theorem   error   data   let    function   log    case   bound   analysis   sample   probability   algorithm   random   distribution   rank   learning   given   problem   linear        & 0.32 \\
Neural network &		time    neurons   neural   neuron   input   spike   network   signal   figure   model   noise   synaptic   information   output   fig    rate   function   state   dynamics   learning         & 0.54 \\
\bottomrule
	\end{tabular*}%
	\caption{\small Top 20 words of the topics estimated by LDA and CDTM.} \label{tab:top20}%
\end{table}%

\section{Conclusion} \label{sec:conclusion}
We propose a Concentrated Document Topic Model(CDTM), which is able to produce concentrated and sparse document-topic distributions. In particular, we add an exponential negative entropy of the document-topic parameters as a productive penalty to the posterior. Entropy is used to measure the topic concentration. Concentrated document-topic distributions have low entropy values and thus are penalized less, while the more uniformly distributed document-topic distributions have high entropy values and thus are penalized more. We then apply our proposed model to the NIPS dataset and observe improved topic coherence scores and more concentrated document-topic distributions. Our proposed model could potentially improve the information retrieval accuracy comparing to LDA.

\begin{appendices}
	\section{Expectation of penalty under the variational distribution}
	In this section, we derive the equation of the expectation of the penalty term under the variational distribution. We first find the equation for $E_q[\theta_i \log \theta_i]$.  
	\begin{equation*}
		\begin{aligned}
			E_q[\theta_i \log \theta_i] &= \int \theta_i \log \theta_i \frac{\Gamma(\sum_{l=1}^{K}  \gamma_i)}{\prod_{l=1}^{K}\Gamma(\gamma_l)}\theta_l ^{\gamma_l - 1} d \theta \\
			 &= \frac{\gamma_i}{\sum_{l=1}^{K} \gamma_l} \int \log \theta_i \frac{\Gamma(\sum_{l=1}^{K} \gamma_l^\prime)}{\prod_{l=1}^{K}\Gamma(\gamma_l^\prime)}\theta_l ^{\gamma_l^\prime - 1} d\theta \\
			 & =  \frac{\gamma_i}{\sum_{l=1}^{K} \gamma_l} [\Psi(\gamma_i^\prime) - \Psi(\sum_{l=1}^{K}\gamma_l^\prime)] \\
			 & = \frac{\gamma_i}{\sum_{l=1}^{K} \gamma_l} [\Psi(\gamma_i) + 1/\gamma_i - \Psi(\sum_{l=1}^{K}\gamma_l) - 1/\sum_{l=1}^{K}\gamma_l] \\
		\end{aligned}
	\end{equation*}
 where in the second equality, $\gamma_i^\prime = \gamma_i + 1$ and $\gamma_j^\prime = \gamma_j, \forall j \ne i$, and we make use of $\Gamma(x+1) = x\Gamma(x)$. The third equality follows from \cite{LDA}. Then the expectation of the penalty term under the varitaional distribution takes following form
	\begin{equation*}
		\begin{aligned}
			E_q[-\lambda H(\theta)] &= \lambda \sum_{i=1}^{K} E_q[\theta_i \log \theta_i] \\
					& = \lambda (\frac{\sum_{i=1}^{K}\gamma_i \Psi(\gamma_i)}{\sum_{i=1}^{K}\gamma_i} - \Psi(\sum_{i=1}^{K}\gamma_i) + \frac{K-1}{\sum_{i=1}^{K}\gamma_i})
		\end{aligned}
	\end{equation*}
\end{appendices}

\bibliography{CDTM_Reference.bib}
\end{document}